\documentclass[letterpaper]{article}
\usepackage[preprint]{aaai2027}
\usepackage[hyphens]{url}  % DO NOT CHANGE THIS
\usepackage{graphicx} % DO NOT CHANGE THIS
\usepackage{natbib}  % DO NOT CHANGE THIS AND DO NOT ADD ANY OPTIONS TO IT
\usepackage{caption} % DO NOT CHANGE THIS AND DO NOT ADD ANY OPTIONS TO IT
\usepackage{algorithm}
\usepackage{algorithmic}
\usepackage{booktabs}

\usepackage{amsfonts,bm}
\usepackage{amsmath}
\usepackage{amssymb}
\usepackage{microtype}
\usepackage{colortbl}
\usepackage{enumitem}
\usepackage{tabularx}
\usepackage{xstring}
\usepackage{multirow}
\usepackage{xspace}
\usepackage{subcaption}
\usepackage{pifont}

\newcommand{\nbf}[1]{{\noindent \textbf{#1.}}}

\newcommand{\ours}{AffordAny\xspace}

\newcommand{\cmark}{\ding{51}}
\newcommand{\xmark}{\ding{55}}

\definecolor{tabhighlight}{HTML}{e5e5e5}
\definecolor{lightCyan}{rgb}{0.925,1,1}

\newcommand{\Ls}{\mathcal{L}}

\DeclareMathAlphabet{\mathsfit}{\encodingdefault}{\sfdefault}{m}{sl}
\SetMathAlphabet{\mathsfit}{bold}{\encodingdefault}{\sfdefault}{bx}{n}

\title{AffordAny: Open-World 3D Affordance Grounding from Monocular RGB Images via Vision-Language-Guided Geometric Reasoning}

\author{
    Junqi Wu\textsuperscript{1}\equalcontrib,
    Kaihua Tang\textsuperscript{1}\equalcontrib,
    Xuanwen Chen\textsuperscript{1},
    Hongzhi Li\textsuperscript{1},
    Jianqiang Huang\textsuperscript{2}\corresponding,
    Xian-Sheng Hua\textsuperscript{1}\corresponding
}
\affiliations{
    \textsuperscript{1}Tongji University, Shanghai, China\\
    \textsuperscript{2}Chinese Academy of Sciences, China
}

\begin{document}

\maketitle

\begin{abstract}
Open-world 3D affordance grounding requires localizing functional object parts in 3D given free-form language queries.
Existing methods typically assume pre-built object-centric 3D geometry and closed affordance ontologies, limiting deployment from raw RGB observations.
We present \ours, an end-to-end framework that uses one monocular RGB image to construct large-scale text-conditioned 3D part supervision, ground affordances with a frozen vision-language model (VLM) guided decoder, and improve open-world generalization through pseudo-label self-training.
Our automated pipeline produces a benchmark of 5,334 objects and 10,633 part-level samples spanning 473 categories, an order-of-magnitude increase in categorical diversity over prior work.
The decoder progressively fuses frozen Cosmos-2B features with 3D geometry through spatial projection, instruction-conditioned semantic compression, and bidirectional geometry-semantics interaction.
Minimal-perturbation pseudo-label self-training further adds new objects without human annotation.
Under a systematic generalization protocol evaluating unseen objects, unseen categories, and unseen instruction paraphrases, our approach achieves 0.428 IoU on unseen objects and 0.315 IoU on unseen categories after self-training, with unseen-category mIoU improving by 6.3\% relative ($p{<}0.01$) and an instruction sensitivity gap of only 0.105, demonstrating effectiveness and robustness of our method. The code\footnote{Code: \url{https://github.com/lzlfwow/AffordAny}.} and dataset\footnote{Dataset: \url{https://modelscope.cn/datasets/lzlfwow/AffordAny}.} are publicly available. 
\end{abstract}

\section{Introduction}
\label{sec:intro}

3D affordance grounding aims to localize functional interaction regions on objects given free-form language or action descriptions~\cite{deng20213d}.
For embodied agents, accurately identifying where and how to interact with objects (the handle of a mug, the lid of a box, the seat of a chair) is a prerequisite for manipulation, navigation, and task planning~\cite{mo2021where2act,luo2022learning}.
In realistic deployments, this problem is inherently open-world: agents encounter long-tail objects, diverse part taxonomies, and unconstrained natural-language instructions, while their observations are typically raw RGB images rather than pre-existing 3D assets~\cite{hong20233dllm,huang2023embodied}.

Recent work has begun to address open-vocabulary affordance grounding by leveraging vision-language representations.
LASO~\cite{li2024laso} proposes a language-conditioned part segmentation approach that uses text-query-based bidirectional fusion for 3D affordance localization.
LMAffordance3D~\cite{zhu2025grounding} introduces affordance triples with CLIP-based spatial features for language-guided 3D part prediction.
OpenAD~\cite{nguyen2023open} builds an open-vocabulary alignment between text embeddings and 3D point features via cosine-similarity matching.
While these methods demonstrate promising results, they share several key limitations.
First, they assume that pre-built 3D geometry is available at both training and inference time, which hides a substantial portion of the real deployment problem.
Second, scalable datasets with real-image provenance and 3D part-level supervision remain scarce~\cite{deng20213d}, particularly when target semantics are expressed by free-form text instructions rather than a fixed closed ontology.
Third, current text-encoding approaches based on CLIP~\cite{radford2021learning} or hashed embeddings are brittle to instruction paraphrasing~\cite{zhu2025grounding}, a critical limitation for practical language-conditioned interaction.

\begin{figure*}[t]
  \centering
  \includegraphics[width=\textwidth]{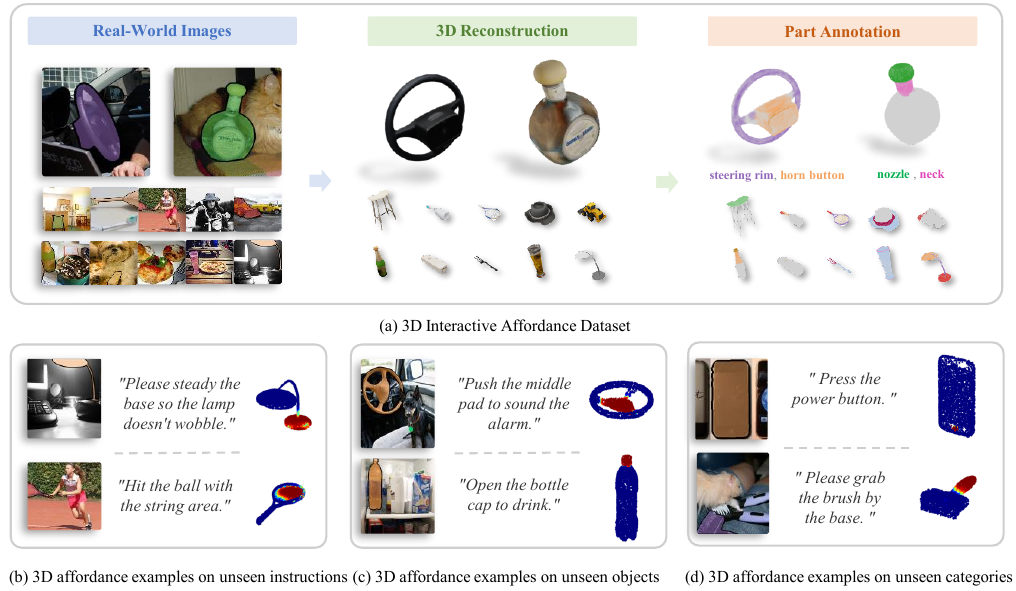}
  \caption{Overview of the proposed dataset. Given an RGB image and instruction, the pipeline reconstructs object-centric 3D geometry and grounds the target part. We evaluate unseen instructions, unseen objects, and unseen categories.}
  \label{fig:teaser}
\end{figure*}

This paper presents Affordance Anything (\ours), an end-to-end framework that bridges these gaps through three complementary contributions.
First, we introduce an automated data construction pipeline that starts from the large-vocabulary LVIS dataset~\cite{gupta2019lvis} and produces text-conditioned 3D part-level affordance supervision through instance filtering, single-image 3D reconstruction with SAM 3D~\cite{chen2025sam}, LLM-guided part prompt generation, and multi-view annotation fusion.
Starting from 1{,}270{,}141 LVIS training instances across 1{,}203 categories, our pipeline retains 473 interaction-relevant categories, selects 7{,}407 candidate objects, and produces a benchmark containing 5{,}334 objects with 10{,}633 part-level samples and 31{,}899 instruction-level samples.
Second, we propose a VLM-guided cross-modal 3D affordance decoder that leverages frozen Cosmos-2B~\cite{agarwal2025cosmos} features to bridge language semantics and 3D geometry.
The decoder extracts a global affordance token and dense visual tokens from the source image, and fuses them with 3D point-cloud features through projection injection, text-conditioned semantic bottleneck, and bidirectional prototype-point fusion, enabling robust language-conditioned part grounding without fine-tuning the VLM backbone.
Third, we introduce a pseudo-label self-training strategy with minimal perturbation training and post-hoc Platt calibration, which leverages the trained decoder to generate affordance labels on new objects and significantly improves category-level generalization without additional human annotation.

We evaluate under a systematic generalization protocol that separately measures performance on unseen objects (same category, new instances), unseen categories (entirely novel object types), and unseen instruction paraphrases (same object and part, different wording).
Notably, our entire pipeline takes only a single monocular RGB image and object masks as input, requiring no pre-built 3D assets or multi-view captures at inference time.
Our decoder achieves 0.428 IoU on unseen objects and 0.305 IoU on unseen categories, while reducing the instruction sensitivity gap to only 0.105.
After pseudo-label self-training, segmentation metrics improve significantly, notably unseen-category mIoU by 6.3\% relative ($p{<}0.01$), with the largest gains on underrepresented tail categories, demonstrating that the framework can bootstrap its own training data to improve open-world generalization.

We make the following contributions:
\begin{itemize}[leftmargin=1.5em]
\item We present the first automated pipeline for constructing large-scale text-conditioned 3D part-level affordance supervision from real images, producing a benchmark spanning 473 categories, 5{,}334 objects, and 10{,}633 part-level samples.
\item We propose a VLM-guided decoder for open-vocabulary affordance grounding that effectively fuses Cosmos-2B features with 3D point-cloud geometry.
%fuses frozen Cosmos-2B features with 3D point-cloud geometry through projection injection, text-conditioned bottleneck, and bidirectional fusion.
\item We introduce a pseudo-label self-training strategy with minimal perturbation training and Platt calibration, significantly improving category-level generalization without additional human annotation. % notably unseen-category mIoU by 6.3\% relative ($p{<}0.01$), .
\item We design a systematic evaluation protocol covering unseen objects, unseen categories, and unseen instructions. Our method demonstrates strong open-world generalization and notably low sensitivity to instruction variations.
\end{itemize}

\section{Related Work}
\label{sec:related}

\noindent\textbf{3D Affordance Learning.}
Affordance reasoning has been studied across 2D perception~\cite{do2018affordancenet}, robotic manipulation~\cite{mo2021where2act}, and 3D scene understanding.
PartNet~\citep{mo2019partnet} provides fine-grained hierarchical part annotations, while SAPIEN~\citep{xiang2020sapien} supports physically grounded interaction with articulated objects.
3D AffordanceNet~\citep{deng20213d} provides probabilistic point-level labels, while AGD20K~\citep{luo2022learning} captures affordance grounding in real interaction images.
AffordPose~\citep{jian2023affordpose} couples part-level 3D affordances with hand poses, and AffordBridge~\citep{vu2026affordmatcher} extends functional-interaction annotations to real 3D scenes.
LMAffordance3D~\citep{zhu2025grounding} further introduces language-conditioned 3D affordance triples, but operates within a closed vocabulary.
In contrast, our work starts from \emph{real images}, constructs supervision \emph{automatically}, and supports \emph{open-vocabulary} text queries.

\noindent\textbf{Vision-Language Models for 3D Understanding.}
ULIP~\citep{xue2023ulip}, OpenShape~\citep{liu2023openshape}, and Uni3D~\citep{zhou2024uni3d} transfer image--text priors to open-world point-cloud representations, while PartSLIP~\citep{liu2023partslip} specializes this transfer for low-shot 3D part segmentation.
At the scene and instruction level, 3D-LLM~\citep{hong20233dllm}, PointLLM~\citep{xu2024pointllm}, and LEO~\citep{huang2023embodied} align 3D inputs with language models for grounding, question answering, or embodied reasoning.
LISA~\citep{lai2024lisa} decodes a special language token into a 2D mask, inspiring our \texttt{<AFF>} token, whereas AffordanceLLM~\citep{qian2024affordancellm} remains image based.
We instead extract frozen VLM features offline and fuse them with point geometry through a lightweight decoder, retaining pretrained alignment without VLM fine-tuning.

\noindent\textbf{Open-Vocabulary Visual Grounding.}
CLIP~\cite{radford2021learning} enables open-vocabulary recognition; Grounding DINO~\cite{liu2024grounding} localizes free-form queries, while ODISE~\cite{xu2023open} and VLPart~\cite{sun2023going} extend language supervision to dense or part-level prediction.
Promptable models such as SAM~\cite{kirillov2023segment} and PixelLM~\cite{ren2024pixellm} further connect free-form prompts or multimodal reasoning to pixel masks.
In 3D, PLA~\cite{ding2023pla}, OpenScene~\cite{peng2023openscene}, and OV3D~\cite{jiang2024open} lift vision-language knowledge to point segmentation, while OpenMask3D~\cite{takmaz2023openmask3d} targets open-vocabulary 3D instances.
Feature 3DGS~\cite{zhou2024feature}, LangSplat~\cite{qin2024langsplat}, and OpenGaussian~\cite{wu2024opengaussian} instead embed open-vocabulary features in Gaussian fields.
Our pipeline uses 2D foundation models only to mine supervision and lift it by multi-view fusion; the final decoder predicts directly on point geometry conditioned on frozen VLM features.

\section{\ours Benchmark}
\label{sec:benchmark}

Prior datasets range from synthetic objects~\cite{deng20213d,jian2023affordpose} to real-image or scene interactions~\cite{luo2022learning,vu2026affordmatcher}, but even language-conditioned methods generally map free-form instructions to fixed part/affordance taxonomies; Table~\ref{tab:dataset_comparison} marks these as closed under \emph{Open Part}.
We introduce an automated pipeline that constructs large-scale, text-conditioned, 3D part-level affordance supervision entirely from real images.
Starting from the LVIS dataset~\cite{gupta2019lvis}, the pipeline produces a benchmark of 5{,}334 objects and 10{,}633 part-level samples spanning 473 categories and 678 unique part types, an order-of-magnitude increase in categorical diversity over prior work.
Figure~\ref{fig:pipeline_overview} gives an overview of the pipeline; full construction details are provided in Appendix.

\begin{figure*}[t]
\centering
\includegraphics[width=0.95\textwidth]{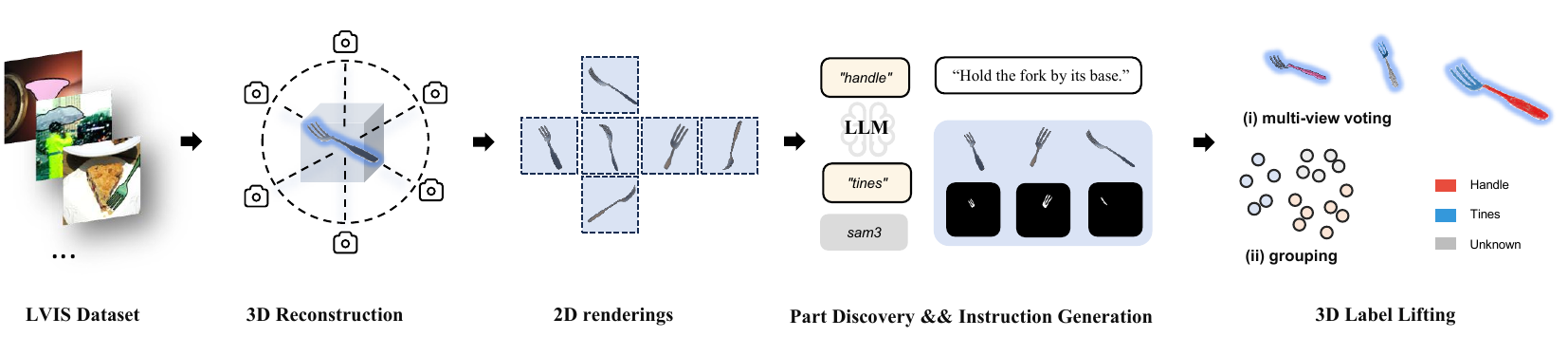}
\caption{\ours data construction pipeline: LVIS images are reconstructed and rendered from six views; an LLM proposes parts and instructions, a segmentation model labels 2D views, and multi-view voting lifts labels to 3D.}
\label{fig:pipeline_overview}
\end{figure*}

\begin{table*}[t]
  \centering
  \small
  \setlength{\tabcolsep}{1mm}
  \begin{tabularx}{\textwidth}{@{}>{\raggedright\arraybackslash}p{1.48in}c>{\raggedright\arraybackslash}X>{\raggedright\arraybackslash}X>{\raggedright\arraybackslash}p{0.70in}ccccc@{}}
  \toprule
  & & & & & \multicolumn{2}{c}{Taxonomy} & \multicolumn{3}{c}{Open} \\
  \cmidrule(lr){6-7} \cmidrule(lr){8-10}
  Dataset & Year & Source / Input & Annotation & Scale & Obj. & Part & Part & Obj. & Cat. \\
  \midrule
  3D AffordanceNet~\citep{deng20213d} & 2021 & CAD / point cloud & 3D heatmap & 23K shapes & 23 & 18 & \xmark & \cmark & \xmark \\
  GAPartNet~\citep{geng2023gapartnet} & 2023 & Articulated CAD / point cloud + mesh & 3D part mask + pose & 8{,}489 parts & 27 & 9 & \xmark & \cmark & \cmark \\
  AffordPose~\citep{jian2023affordpose} & 2023 & CAD / mesh + affordance & 3D part + hand pose & 26.7K inter. & 13 & 8 & \xmark & \xmark & \xmark \\
  LASO~\citep{li2024laso} & 2024 & CAD / point cloud + language & 3D part mask & 19{,}751 pairs & 23 & 17 & \xmark & \cmark & \cmark \\
  AGPIL~\citep{zhu2025grounding} & 2025 & CAD + image / image + point cloud + language & 3D heatmap & 30{,}972 trip. & 23 & 17 & \xmark & \cmark & \cmark \\
  3DAffordSplat~\citep{wei20253daffordsplat} & 2025 & 3DGS / point cloud + 3DGS & 3D heatmap + 3DGS & 6{,}631 labels & 21 & 18 & \xmark & \cmark & \cmark \\
  \midrule
  SceneFun3D~\citep{delitzas2024scenefun3d} & 2024 & Real scan / scene + language & 3D part + motion & 14.8K ann. & -- & 9 & \xmark & \cmark & \xmark \\
  AffordBridge~\citep{vu2026affordmatcher} & 2026 & Real scan / scene + image & 3D interaction labels & 291{,}637 ann. & 157 & -- & \xmark & -- & -- \\
  \midrule
  \textbf{\ours (Ours)} & 2026 & Real image / image & 3D heatmap + 3DGS + text & 31{,}899 pairs & \textbf{473} & \textbf{678} & \cmark & \cmark & \cmark \\
  \bottomrule
  \end{tabularx}
  \caption{Comparison with related affordance and interactive-region datasets. Under ``Open,'' Part denotes an unrestricted part-level vocabulary, while Obj.\ and Cat.\ denote evaluation on unseen instances
  and categories; dashes indicate unavailable results.}
  \label{tab:dataset_comparison}
  \end{table*}

\begin{figure}[t]
\centering
\includegraphics[width=\columnwidth]{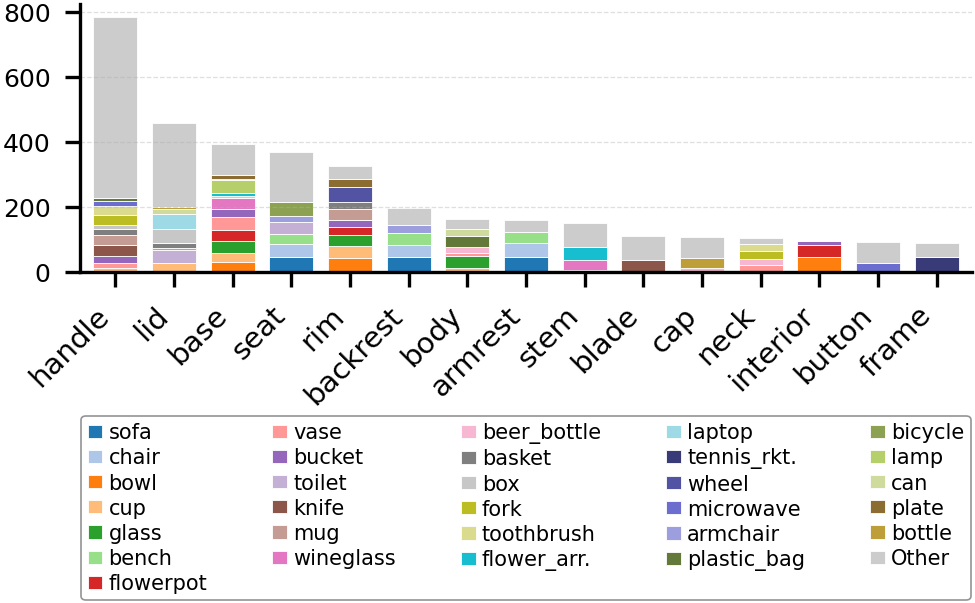}
\caption{Part--category distribution for the most frequent parts and categories.}
\label{fig:dataset_stats}
\end{figure}

\nbf{Pipeline overview}
The pipeline begins with category-level and instance-level filtering of LVIS training instances, retaining 7{,}407 candidate objects from 473 interaction-relevant categories.
Each object undergoes single-image 3D reconstruction with SAM 3D~\cite{chen2025sam} into 3D Gaussians~\cite{kerbl20233d}, multi-view rendering from six canonical viewpoints, LLM-guided part discovery, text-conditioned 2D part segmentation, and multi-view 3D label lifting.
For each validated part, three synonymous natural-language interaction instructions are generated, yielding 31{,}899 instruction-level samples.

\nbf{Evaluation protocol}
We adopt a multi-axis generalization protocol: 70\% of categories are designated as seen (training), with 15\%/15\% held out for validation/testing of unseen-category generalization; within seen categories, objects are split 80\%/10\%/10\% for unseen-object evaluation; two of three synonymous instructions are used for training, with the third reserved for instruction-robustness evaluation.
Detailed split statistics are provided in Appendix.

\section{Method}
\label{sec:method}

Given an RGB image $I$, an object mask $M$, and a free-form instruction $l$. Our framework first reconstructs an object-centric point cloud $\mathbf{P} = \{\mathbf{p}_i\}_{i=1}^{N}$ using SAM 3D~\cite{chen2025sam}, \textit{i.e.}, $P=\mathrm{SAM3D}(I,M)$. The task is then to predict point-wise scores $\hat{s} \in [0,1]^N$ for the functional region specified by $l$.

\subsection{VLM-Guided 3D Affordance Decoder}
\label{sec:vlm_decoder}

\begin{figure*}[t]
\centering
\includegraphics[width=0.95\textwidth]{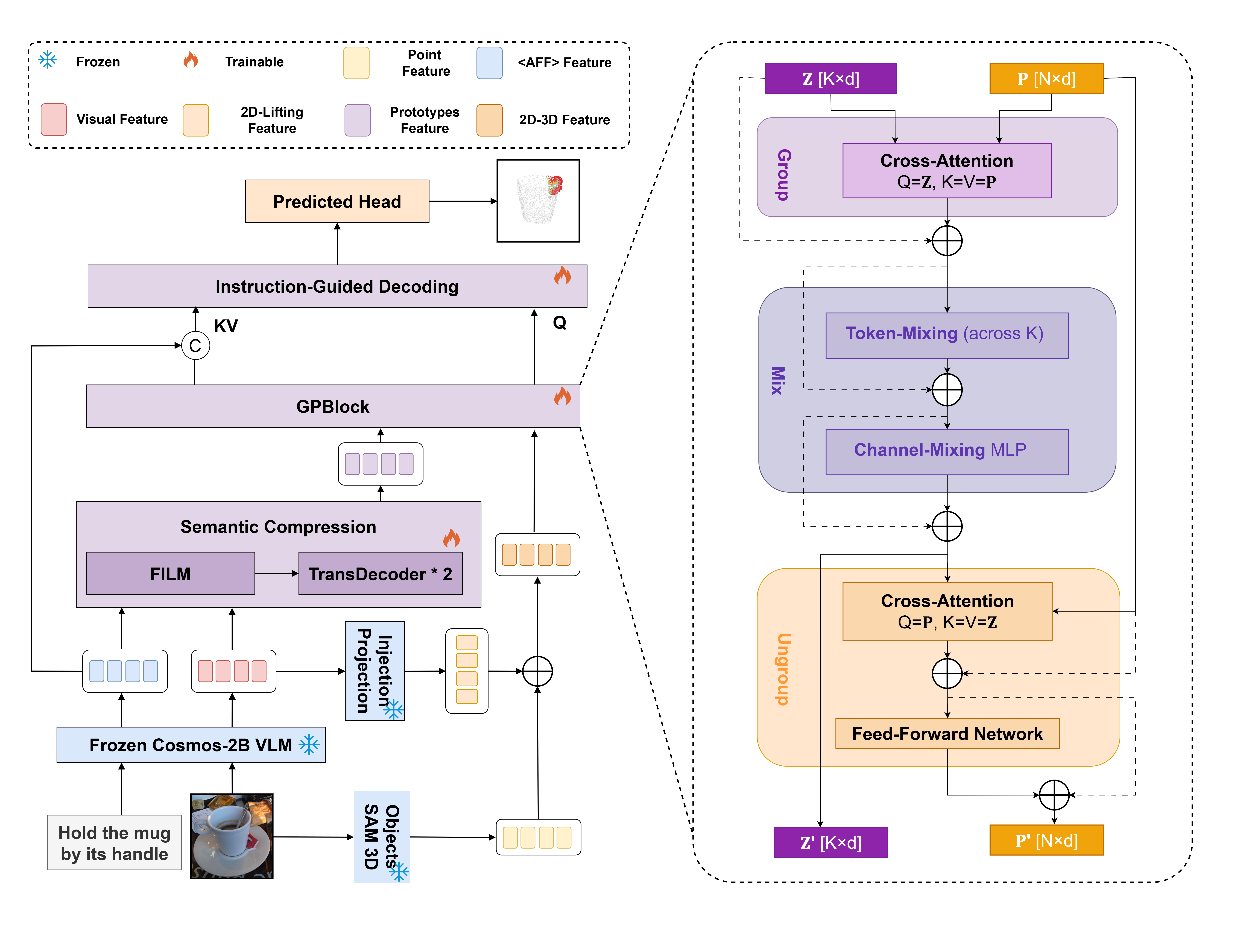}
\caption{Architecture overview of \ours. A frozen Cosmos-2B VLM encodes the instruction and image; projection injection, semantic compression, and GPBlock bidirectional fusion combine these features with 3D geometry to predict per-point affordance scores.}
\label{fig:architecture}
\end{figure*}

Figure~\ref{fig:architecture} shows three fusion stages: projection injection supplies local image evidence, semantic compression distills instruction-relevant VLM features, and bidirectional attention grounds them in 3D geometry.
The underlying principle is that image--language evidence identifies the functional part, whereas geometry constrains its spatial extent.

\nbf{Offline VLM Feature Extraction}
We run a frozen Cosmos-2B VLM on the masked image and an instruction augmented with an \texttt{<AFF>} marker. Its contextual marker feature $\mathbf{h}_{\text{aff}} \in \mathbb{R}^{D}$ summarizes image--instruction semantics, while $144$ visual tokens $\mathbf{F}_v \in \mathbb{R}^{144 \times D}$ retain spatial layout ($D=2048$). Both are cached offline; the VLM and marker embedding remain frozen.
Thus, the marker provides a query-specific summary and the visual tokens preserve local image evidence. Freezing also makes the costly cross-modal encoder reusable across decoder variants.
It further avoids adapting the backbone specifically to the seen-category distribution.

\nbf{Point Encoder}
Point features $\mathbf{F} \in \mathbb{R}^{N \times 13}$ and coordinates $\mathbf{X}$ are projected to $d=256$, summed, and refined by residual MLPs into point tokens $\mathbf{P}$. The $145$ VLM tokens $[\mathbf{h}_{\text{aff}};\mathbf{F}_v]$ are projected to the same dimension.
Separate encoding preserves geometric cues until the fusion stages.

\nbf{Spatial Correspondence via Projection Injection}
Using the reconstruction camera, we project each 3D point to $(u,v)$, bilinearly sample the reshaped $12\times12$ grid of $\mathbf{F}_v$, and add the local feature to its point token:
\begin{equation}
\mathbf{P} \leftarrow \mathbf{P} + \text{MLP}\bigl(\text{BilinearSample}(\mathbf{F}_v, \; u, v)\bigr).
\end{equation}
This gives thin or partly visible regions direct access to image evidence at their projected locations.
In contrast to global pooling, the injection preserves the reconstruction-derived correspondence between individual 3D points and source-image regions.

\nbf{Instruction-Conditioned Semantic Compression}
We compress the $145$ VLM tokens into $K=16$ instruction-conditioned prototypes. Learnable embeddings $\{\mathbf{z}_k\}$ are modulated by $\mathbf{h}_{\text{aff}}$ with FiLM~\cite{perez2018film}:
\begin{equation}
\tilde{\mathbf{z}}_k = \mathbf{z}_k \odot (1 + \boldsymbol{\gamma}) + \boldsymbol{\beta}, \quad
\boldsymbol{\gamma} = W_\gamma \mathbf{h}_{\text{aff}}, \;
\boldsymbol{\beta} = W_\beta \mathbf{h}_{\text{aff}},
\end{equation}
where $W_\gamma, W_\beta \in \mathbb{R}^{d \times D}$ are learnable projections from the frozen VLM dimension $D$ to the decoder dimension $d$.
The conditioned prototypes attend to the VLM sequence with a 2-layer Transformer decoder:
\begin{equation}
\mathbf{Z} = \text{TransDec}(\text{Q} = \tilde{\mathbf{Z}}, \; \text{KV} = [\mathbf{h}_{\text{aff}}; \mathbf{F}_v]) \in \mathbb{R}^{K \times d}.
\end{equation}
This compact representation bounds subsequent point--prototype attention independently of the VLM sequence length.
The prototypes therefore form a task-aware semantic bottleneck rather than a generic global pooling of VLM tokens.
FiLM makes this fixed-size bottleneck instruction-dependent before it reads the full VLM sequence.

\nbf{GPBlock: Bidirectional Geometry-Semantics Interaction}
GPBlock (Group-Mix-Ungroup) makes the two representations mutually informative:
\begin{enumerate}[leftmargin=1.5em]
\item \textbf{Group:} Prototypes attend to points ($\text{Q}{=}\mathbf{Z}$, $\text{KV}{=}\mathbf{P}$) to absorb 3D context.
\item \textbf{Mix:} Token- and FFN-based channel-mixing coordinate the prototypes.
\item \textbf{Ungroup:} Points attend to refined prototypes ($\text{Q}{=}\mathbf{P}$, $\text{KV}{=}\mathbf{Z}$), followed by an FFN.
\end{enumerate}
This produces geometry-aware prototypes $\mathbf{Z}'$ and enriched points $\mathbf{P}'$: Group grounds semantics in shape, and Ungroup returns the grounded context to each point.
Such bidirectionality supports transfer when one instruction refers to different geometric realizations across categories.

\nbf{Instruction-Guided Decoding}
The enriched point tokens query the affordance token and refined prototypes through a four-layer Transformer cross-decoder:
\begin{equation}
\mathbf{H} = \text{TransDec}(\text{Q}=\mathbf{P}',\; \text{KV}=[\mathbf{h}_{\text{aff}};\mathbf{Z}']).
\end{equation}
The decoded tokens are mapped to per-point logits by a lightweight head, with sigmoid applied at inference:
\begin{equation}
\hat{s}_i = \sigma\!\bigl(\text{Head}(\mathbf{H}_i)\bigr).
\end{equation}
This final cross-attention lets every point retrieve both the global instruction cue and its geometry-refined semantic prototypes.

\nbf{Instruction Dropout}
During training, instruction dropout randomly zeros $\mathbf{h}_{\text{aff}}$ and $\mathbf{F}_v$ to discourage semantic shortcuts and strengthen geometric priors.
It also improves robustness when test instructions differ lexically from the training queries.

\subsection{Training}
\label{sec:training}

The model is trained with a combination of three losses:
\begin{equation}
\Ls = \lambda_{\text{bce}} \Ls_{\text{bce}} + \lambda_{\text{focal}} \Ls_{\text{focal}} + \lambda_{\text{dice}} \Ls_{\text{dice}},
\end{equation}
where weighted BCE, focal~\cite{lin2017focal}, and Dice address class imbalance, hard points, and region overlap, respectively. Appendix gives full settings.
BCE preserves point-wise calibration, focal loss emphasizes difficult minority points, and Dice directly optimizes sparse-region overlap.

\subsection{Pseudo-Label Self-Training}
\label{sec:selftraining_method}

We use the trained decoder as a teacher and calibrate its predictions post hoc, separating segmentation changes from probability calibration.

\nbf{Pseudo-label generation}
We process 5{,}325 disjoint LVIS objects, ignore probabilities in $[0.15,0.65]$, binarize the rest, and subsample pseudo-labels to a 1:1 ratio with real data.
The ignored band retains only confident foreground and background, while subsampling prevents noisy pseudo-labels from dominating ground-truth supervision.

\nbf{Minimal perturbation training}
Starting from the supervised model, we run one additional round with pseudo-label weight $\bar{w}=0.1$ and an extra $0.5\times$ BCE/Focal scaling (about $5\%$ effective weight).
This conservative update exposes the decoder to additional intra-category geometry while limiting drift in its supervised representation and probability calibration.
In preliminary experiments, more aggressive mixing improved thresholded overlap but distorted predicted probabilities; we therefore use one low-weight round.

\nbf{Post-hoc Platt calibration}
To correct the teacher's positive-rate bias, we apply Platt scaling~\cite{platt1999probabilistic} after training:
\begin{equation}
\hat{s}_i^{\text{cal}} = \sigma(a \cdot z_i + b),
\end{equation}
Here $z_i$ is the pre-sigmoid logit and validation selects $(a,b)$. The selected $(1.10,-0.75)$ restores MAE but partly attenuates mIoU gains; AUC is unchanged by this monotonic transform. We report both raw and calibrated predictions.
This separates the overlap--calibration trade-off from gains obtained by threshold tuning.
The scale $a$ adjusts sharpness, whereas the offset $b$ corrects the teacher's positive-rate bias.

\section{Experiments}
\label{sec:exp}

\subsection{Evaluation Protocol}

\nbf{Generalization protocol}
Six splits isolate three generalization axes: unseen objects from seen categories (789 test / 710 val), unseen categories (1{,}552 test / 1{,}515 val), and unseen paraphrased instructions on matched object--part pairs. Each part has three synonymous instructions: two for training (12{,}134 samples) and one for robustness evaluation (6{,}067 samples). This separation prevents instance novelty, category novelty, and wording changes from being conflated in one score.
For instruction robustness, matching the object and target part ensures that only wording changes between the paired evaluations.

\nbf{Metrics}
We report pooled point-wise IoU at threshold 0.5 and category-macro mIoU (intersection and union are pooled per category before averaging). We also report global point-wise ROC AUC, point-wise MAE, and SIM: mean per-sample histogram intersection after unit-mass normalization of predictions and soft targets.
The macro score gives each category equal weight, while pooled IoU reflects aggregate point-level performance; reporting both exposes gains dominated by frequent categories.

\subsection{Implementation Details}
Our decoder has width $d=256$, 16 prototypes, one GPBlock, and a four-layer cross-decoder. Frozen Cosmos-2B features are cached offline. We train for 300 epochs with AdamW, cosine decay, mixed-precision DDP on eight GPUs, and select by validation IoU. Appendix gives full settings.
The supervised objective combines weighted BCE, focal, and Dice losses, with instruction dropout used only during training. Baselines use the frozen visual and language features prescribed by their designs.
No VLM parameters receive gradients during decoder training; all reported improvements therefore arise from the decoder and its fusion of cached features with geometry.

\subsection{Baselines}

We compare OpenAD (CLIP text alignment), LASO (GPBlock and text-query RPD)~\cite{li2024laso}, LMAffordance3D (CLIP token fusion)~\cite{zhu2025grounding}, and \ours. All use identical point clouds, masks, splits, losses, 300-epoch budgets, and checkpoint selection; method-specific features and decoder capacities are not matched.
Consequently, the comparison controls data and supervision rather than parameter count or feature-extraction cost.

\subsection{Main Results}

\begin{table*}[t]
\centering
\small
\setlength{\tabcolsep}{2.5pt}
\begin{tabular}{lc|ccccc|ccccc|ccccc}
\toprule
\multirow{2}{*}{Model} & \multirow{2}{*}{\shortstack{Best\\Val}}
& \multicolumn{5}{c|}{Unseen Instruction}
& \multicolumn{5}{c|}{Unseen Object}
& \multicolumn{5}{c}{Unseen Category} \\
\cmidrule(lr){3-7} \cmidrule(lr){8-12} \cmidrule(lr){13-17}
& & IoU & mIoU & AUC & MAE$\downarrow$ & SIM
& IoU & mIoU & AUC & MAE$\downarrow$ & SIM
& IoU & mIoU & AUC & MAE$\downarrow$ & SIM \\
\midrule
OpenAD & .413
& .484 & .390 & .878 & .255 & .360
& .362 & .272 & .783 & .304 & .294
& .266 & .234 & .683 & .345 & .257 \\
LASO & .453
& .451 & .320 & .853 & .205 & .374
& .394 & .282 & .820 & .222 & .324
& .266 & .227 & \textbf{.723} & \textbf{.264} & .262 \\
LMAffordance3D & .454
& .469 & .330 & .852 & .188 & .387
& .418 & .301 & .820 & \textbf{.203} & .323
& .244 & .211 & .669 & .278 & .223 \\
\ours (Ours) & \textbf{.473}
& \textbf{.680} & \textbf{.630} & \textbf{.956} & \textbf{.112} & \textbf{.604}
& \textbf{.428} & \textbf{.306} & \textbf{.829} & .212 & \textbf{.339}
& \textbf{.305} & \textbf{.243} & .716 & .266 & \textbf{.268} \\
\bottomrule
\end{tabular}
\caption{Decoder comparison after 300 epochs. ``Unseen Instruction'' uses held-out paraphrases. Bold is best; $\downarrow$ is lower-is-better.}
\label{tab:decoder_comparison_300ep}
\end{table*}

Table~\ref{tab:decoder_comparison_300ep} shows that \ours leads in IoU on unseen instructions (0.680), unseen objects (0.428 vs.~0.418), and unseen categories (0.305 vs.~0.266). It also has the best unseen-object AUC and SIM and unseen-category SIM. LASO retains the best unseen-category AUC and MAE, distinguishing overlap and concentration from ranking and calibration.
The protocol also exposes baseline-specific failures: LMAffordance3D is competitive on unseen objects but falls to 0.244 IoU on unseen categories.
We therefore do not claim uniform dominance: the main gain is better thresholded localization and spatial concentration under category shift.

\subsection{Instruction Robustness}
\label{sec:instruction_robustness}

We compare seen instructions with held-out paraphrases on matched object--part pairs, e.g., ``grab the handle'' versus ``reach for the grip.''

\begin{table}[!htbp]
\centering
\small
\setlength{\tabcolsep}{3pt}
\begin{tabular}{lccccc}
\toprule
Model & $\Delta$IoU & $\Delta$mIoU & $\Delta$AUC & $\Delta$MAE & $\Delta$SIM \\
\midrule
OpenAD & -0.166 & -0.166 & -0.087 & 0.075 & \textbf{-0.065} \\
LASO & -0.342 & -0.429 & -0.135 & 0.133 & -0.323 \\
LMAffordance3D & -0.360 & -0.466 & -0.140 & 0.129 & -0.351 \\
\textbf{\ours} & \textbf{-0.105} & \textbf{-0.108} & \textbf{-0.031} & \textbf{0.037} & -0.078 \\
\bottomrule
\end{tabular}
\caption{Instruction robustness on matched object--part pairs (unseen $-$ seen). Smaller absolute gaps are better.}
\label{tab:instruction_robustness}
\end{table}

Table~\ref{tab:instruction_robustness} reports unseen-minus-seen differences. \ours has the smallest absolute IoU, mIoU, AUC, and MAE gaps, including $|\Delta\text{IoU}|=0.105$ versus 0.166 for OpenAD and $>0.34$ for LASO and LMAffordance3D. OpenAD has a slightly smaller SIM gap (0.065 vs.~0.078).
The advantage across overlap, ranking, and error metrics is therefore not attributable to one favorable decision threshold.

\subsection{Ablation Study}

\begin{table}[!htbp]
\centering
\small
\setlength{\tabcolsep}{2.5pt}
\begin{tabular}{lcccccc}
\toprule
\multirow{2}{*}{Configuration} & \multicolumn{3}{c}{Unseen Object} & \multicolumn{3}{c}{Unseen Category} \\
\cmidrule(lr){2-4} \cmidrule(lr){5-7}
 & IoU & AUC & MAE & IoU & AUC & MAE \\
\midrule
\textbf{Full model} & \textbf{.428} & \textbf{.829} & .212 & \textbf{.305} & \textbf{.716} & \textbf{.266} \\
w/o Proj.\ Inj. & .416 & .803 & \textbf{.211} & .275 & .691 & .277 \\
w/o Sem.\ Compr. & .398 & .793 & .219 & .245 & .653 & .276 \\
w/o GPBlock & .403 & .808 & .218 & .270 & .686 & .273 \\
w/o Instr.\ dropout & .402 & .807 & .217 & .288 & .694 & .273 \\
\bottomrule
\end{tabular}
\caption{Ablation of \ours (200 epochs). Semantic compression, GPBlock, projection injection, and instruction dropout contribute $0.060$, $0.035$, $0.030$, and $0.017$ unseen-category IoU, respectively.}
\label{tab:ablation}
\end{table}

Table~\ref{tab:ablation} shows the largest unseen-category IoU drop without semantic compression ($-$0.060), followed by GPBlock ($-$0.035), projection injection ($-$0.030), and instruction dropout ($-$0.017). Thus, category transfer requires an instruction-specific bottleneck plus interaction with local 3D evidence.
Removing bidirectional fusion weakens shape grounding, whereas removing projection injection mainly affects unseen instances as local appearance is decisive.
The smaller but consistent dropout gain supports its role as a regularizer rather than a primary source of semantic knowledge.

\subsection{Pseudo-Label Self-Training}
\label{sec:selftraining}

We use \ours as its own teacher improves unseen-category generalization without additional manual labels.

\nbf{Protocol}
We use 5{,}325 disjoint LVIS objects, ignore predictions with $0.15\le p\le0.65$, and mix a 1:1 pseudo/real subset in one low-weight round ($\bar w=0.1$, extra $0.5\times$ BCE/Focal scaling). We evaluate before and after validation-selected Platt scaling $(a,b)=(1.10,-0.75)$; significance uses paired bootstrap ($B=10{,}000$).
The confidence band, mixing weight, and calibration parameters are fixed before test evaluation.

\begin{table}[!htbp]
\centering
\small
\setlength{\tabcolsep}{3pt}
\begin{tabular}{llcccc}
\toprule
Split & Configuration & IoU & mIoU & AUC & MAE \\
\midrule
\multirow{3}{*}{\shortstack[l]{Unseen\\Object}}
  & Baseline & .428 & .306 & .829 & .212 \\
  & + Self-Training & .428 & \textbf{.315}$^\ddagger$ & \textbf{.832}$^\ddagger$ & .223 \\
  & \quad + Platt cal. & .428 & .306 & \textbf{.832}$^\ddagger$ & \textbf{.211} \\
\midrule
\multirow{3}{*}{\shortstack[l]{Unseen\\Category}}
  & Baseline & .305 & .243 & .716 & .266 \\
  & + Self-Training & \textbf{.315}$^\ddagger$ & \textbf{.259}$^\ddagger$ & \textbf{.720}$^\ddagger$ & .275 \\
  & \quad + Platt cal. & .309 & .246 & \textbf{.720}$^\ddagger$ & \textbf{.265}$^*$ \\
\bottomrule
\end{tabular}
\caption{Pseudo-label self-training before/after Platt calibration. Pre-calibration improves segmentation, especially unseen-category mIoU (+0.016, +6.3\%), while calibration restores MAE but attenuates the gain. $^\ddagger p{<}0.01$; $^* p{<}0.05$.}
\label{tab:selftraining}
\end{table}

\nbf{Results}
Table~\ref{tab:selftraining} shows significant pre-calibration gains in unseen-category mIoU (+0.016, +6.3\% relative), IoU (+0.010), and AUC (+0.004), all $p<0.01$. Platt scaling restores MAE but partly reduces overlap gains while preserving the AUC improvement.
This behavior is expected: scaling changes probability levels but, as a monotonic transform, preserves rank order.

Gains are larger for tail categories ($+0.012$ for $\le50$ training samples, 64/135 improved, vs.~$+0.003$ for head categories); 49/71 unseen categories improve. The raw model is under-confident on rare positives, while Platt scaling improves its reliability curve. We report raw and calibrated results because overlap and probability quality capture different behavior.
This concentration on tail categories supports the intended role of pseudo-labels: adding geometric diversity without new manual annotations.
It does not expand the language vocabulary, so these gains isolate geometric coverage rather than new linguistic supervision.

\subsection{Qualitative Results}

Appendix shows more focused localization than the baselines: \ours preserves target parts under paraphrases and concentrates on functional regions of unseen categories.
In contrast, baseline predictions often become diffuse under paraphrasing.
These examples agree with the quantitative instruction-gap and unseen-category results.

\subsection{Limitations}

Of 7{,}398 completed objects, 2{,}064 have no validated parts, conflating genuinely non-decomposable objects with reconstruction or prompt failures. Long-tailed category and part distributions may favor common parts and weaken estimates for rare ones. Automatically fused multi-view labels also lack human verification and therefore contain unavoidable projection and segmentation noise. Finally, frozen offline VLM features limit task adaptation; end-to-end fine-tuning may improve results but requires substantially more compute.
These factors limit how directly benchmark performance transfers to safety-critical manipulation without further validation.Results on a manually verified test set are provided in Appendix.

\FloatBarrier

\section{Conclusion}
\label{sec:conclusion}

We presented \ours, an end-to-end framework for open-world 3D affordance grounding from a single monocular RGB image.
Our automated pipeline constructs a benchmark of 5{,}334 objects spanning 473 categories with open-vocabulary text-conditioned part supervision from real images.
The VLM-guided decoder fuses frozen Cosmos-2B features with 3D geometry through projection injection, semantic compression, and bidirectional fusion, achieving strong generalization to unseen objects and categories with notably small instruction sensitivity.
Pseudo-label self-training further improves category-level generalization, notably unseen-category mIoU by 6.3\% relative ($p{<}0.01$), with the largest gains on tail categories.
Future work includes end-to-end VLM fine-tuning, scaling the pseudo-label pipeline to larger image sources, and deploying the system for downstream robotic manipulation.

\section{Acknowledgement}
\label{sec:acknowledgement}

This work was supported by the Fundamental Research Funds for the Central Universities at Tongji University under Grant No.~22120260376.

\bibliography{cite}

\clearpage
\appendix
\setcounter{secnumdepth}{2}
\section*{Supplementary Material}
\noindent\textbf{Roadmap.}
This document is organized around the evidence needed to assess the paper.
Appendices~\ref{app:dataset}--\ref{app:dataset_stats} document benchmark construction, quality controls, and split semantics.
Appendices~\ref{app:impl_details}--\ref{app:protocol} specify the decoder, self-training procedure, computational cost, metrics, and statistical protocol.
Appendices~\ref{app:quantitative}--\ref{app:qualitative} provide distributional and qualitative evidence.
Appendix~\ref{app:limitations} summarizes robustness considerations, release constraints, and the scope of the current evaluation.

\section{Dataset Construction and Quality Controls}
\label{app:dataset}

\subsection{Construction Stages and Retained Scale}

The benchmark starts from 1{,}270{,}141 LVIS training instances in 1{,}203 categories.
The successive quality-control stages and retained benchmark scale are summarized in Table~\ref{tab:construction_flow}.
Category filtering first retains 677 categories with a stable operable-part concept.
Image-level filtering and the square-root category quota then produce 7{,}407 candidate instances in 473 categories.
Of these candidates, 7{,}398 reach completed object-level processing.
After reconstruction and part-level quality control, 5{,}334 objects with 10{,}633 validated parts are retained for the benchmark.
The distinction between a \emph{candidate}, a \emph{completed object}, and a \emph{benchmark object} explains the counts used in the main paper.

\begin{table}[H]
\centering
\small
\setlength{\tabcolsep}{4pt}
\begin{tabular}{lrr}
\toprule
Stage & Instances/objects & Categories \\
\midrule
LVIS training instances & 1{,}270{,}141 & 1{,}203 \\
Interactive-category pass & -- & 677 \\
Image-filtered quota candidates & 7{,}407 & 473 \\
Completed object processing & 7{,}398 & 473 \\
Objects with validated parts & 5{,}334 & 473 \\
Validated part samples & 10{,}633 & 473 \\
Generated instruction pairs & 31{,}899 & 473 \\
\bottomrule
\end{tabular}
\caption{Construction flow. Dashes denote a stage for which only the category-level count is used. The experimental manifests use a subset of the generated instruction pairs, as detailed in Table~\ref{tab:dataset_splits_app}.}
\label{tab:construction_flow}
\end{table}

\subsection{Candidate Mining}

Category filtering uses a structured language-model decision that asks whether a category normally contains a meaningful human-interaction or operable region.
Instance filtering uses the masked object as primary evidence and the original image as weak context.
The decision fields test whether the masked instance is recognizable, clear rather than blurry, not severely cropped, and interactive.
The filtering model is \path{gemini-3.1-flash-lite-preview}, with JSON-constrained output, a 60-second timeout, and up to five retries.
Before language-model filtering, deterministic geometry checks require a mask-area ratio of at least 0.03, an aspect ratio no larger than 4.0, and contact with at most one image border (using a 0.02 image-size margin, lower-bounded by two pixels).

To limit category imbalance, the target quota for category $c$ is proportional to $n_c^{0.5}$, where $n_c$ is its number of eligible LVIS images, with at least one candidate per retained category.
After assigning this minimum, integer residual slots are distributed by the largest-remainder rule: floor the square-root-proportional allocation, then rank categories by fractional remainder, weight, and category name.
This operation reduces head-category dominance while preserving naturally diverse category frequencies.

\subsection{Reconstruction and Multi-View Rendering}

Each masked RGB instance is reconstructed with the local SAM-3D-Objects pipeline using its \texttt{hf} checkpoint configuration and random seed 42.
The output includes an object-centric 3D Gaussian representation and camera parameters inferred from the reconstructed point map.
The Gaussian object is rendered at $512\times512$ resolution from six canonical views: front, right, back, left, top, and bottom.
The four horizontal views use yaw angles $0^\circ$, $90^\circ$, $180^\circ$, and $-90^\circ$; the top and bottom views use pitches $89^\circ$ and $-89^\circ$.
All views use radius 2.0 and a $40^\circ$ field of view.
A fixed pre-rotation aligns the source $+z$ direction with the normalized $(1,1,1)$ diagonal before these cameras are applied.

\subsection{Part Discovery and 2D Segmentation}

The recorded benchmark run uses \path{gemini-3.1-flash-lite-preview} for part discovery.
The model proposes at most five candidates in structured JSON and canonicalizes each prompt to ``the \{part\} of the \{object\}.''
Downstream segmentation retains at most three successfully localized parts per object.
The full prompt template requires a visually localizable physical part, rejects whole-object and abstract affordance descriptions, and deduplicates normalized part names.

The six rendered images are segmented with the local SAM3 image model.
The processor confidence threshold is 0.5, the retained-mask score floor is 0.1, and at most three masks are kept per part.
Parts without a successful mask in any view are removed.
These controls ensure multi-view visual support for retained labels, while the human-verified subsets in Appendix~\ref{app:human_review} provide complementary semantic validation.

\subsection{2D-to-3D Lifting and Fusion}

For each view, Gaussian centers are projected with the saved camera parameters.
Depth ownership resolves which Gaussian is visible at each projected pixel, and a Gaussian receives a positive vote when its owned footprint overlaps the 2D part mask.
The footprint radius is $1.75$ times the projected Gaussian scale, capped at 12 pixels; the depth tolerance is the larger of $0.75$ times that scale and $10^{-3}$, and a footprint requires at least 0.02 mask coverage to vote positive.
The per-Gaussian score is the number of positive observations divided by the number of views in which that Gaussian is visible.

Fusion uses a score threshold of 0.3 and an ambiguity margin of 0.05.
Duplicate parts are pruned when their overlap over the smaller mask exceeds 0.5 and their positive-point counts differ by at most 0.3 in relative terms.
Unobserved or low-confidence regions are initially unknown.
Hole filling uses $K=32$ neighbors, requires at least two supporting neighbors and a support ratio of 0.3, permits at most 32 iterations, and caps propagation distance at seven times the median nearest-neighbor spacing.
Points that remain invisible or below threshold after propagation stay unknown and are excluded from packaged supervision.

\subsection{Instruction Generation}

For every packaged part, the recorded benchmark run uses \path{gemini-3.1-flash-lite-preview} with a 120-second timeout and receives the masked source image, object name, part name, and canonical grounding prompt.
It returns exactly three English commands, each limited to 14 words.
The prompt asks for natural human-to-robot instructions whose correct contact or manipulation region is the listed part; it also requests lexical and syntactic variation while prohibiting explicit dataset labels in the spoken command.
Deterministic normalization removes list markers, duplicate strings, and malformed whitespace.
This process yields 31{,}899 generated instruction pairs.

\subsection{Automated Quality Controls and Provenance}

The pipeline records structured rejection reasons, model names, camera metadata, per-view masks, visibility counts, part votes, unknown masks, and package manifests.
This provenance permits stage-wise inspection and deterministic rebuilding of accepted samples.
Together, scalable automatic annotation, deterministic quality controls, and structured provenance support reproducible benchmark construction.
Unknown regions are excluded from packaged supervision to preserve label reliability, while the human-verified subsets in Appendix~\ref{app:human_review} provide independent ground-truth evaluation.

\section{Dataset Statistics and Split Semantics}
\label{app:dataset_stats}

\begin{table}[t]
\centering
\scriptsize
\setlength{\tabcolsep}{1.5pt}
\begin{tabular}{llrrrr}
\toprule
Object split & Instruction use & Inst. rows & Objects & Cats. & Unique parts \\
\midrule
\multirow{2}{*}{Training}
  & Used for optimization & 12{,}134 & 3{,}049 & 331 & 511 \\
  & Unseen instruction   & 6{,}067  & 3{,}049 & 331 & 511 \\
\midrule
\multirow{2}{*}{Validation}
  & Unseen object   & 710   & 351 & 159 & 233 \\
  & Unseen category & 1{,}515 & 764 & 71 & 138 \\
\midrule
\multirow{2}{*}{Test}
  & Unseen object   & 789   & 391 & 209 & 275 \\
  & Unseen category & 1{,}552 & 779 & 71 & 140 \\
\midrule
\multicolumn{2}{l}{\textbf{Experimental manifest total}} & \textbf{22{,}767} & \textbf{5{,}334} & \textbf{473} & \textbf{678} \\
\bottomrule
\end{tabular}
\caption{Experimental split statistics. ``Inst. rows'' counts object--part--instruction records; ``Unique parts'' counts normalized part-name types. For each training object--part pair, instruction indices 1 and 2 are used for optimization and index 0 is reserved for matched unseen-instruction evaluation. Validation and both test protocols also use only instruction index 0. For historical reasons, the optimization and unseen-instruction manifests are named \texttt{train\_unseen\_instruction} and \texttt{train\_seen\_instruction}, respectively. The final row reports unique objects, categories, and part-name types across manifests rather than summing those columns.}
\label{tab:dataset_splits_app}
\end{table}

The benchmark contains 5{,}334 objects, 10{,}633 validated part samples, 678 normalized part names, and 31{,}899 generated instructions.
The experimental manifests contain 22{,}767 rows because they do not evaluate all three generated instructions on every validation and test part.
Each model input contains $N=4{,}096$ sampled Gaussian points with 13 cached geometric/appearance features per point.

The category split assigns 70\% of categories to training and 15\%/15\% to unseen-category validation/test partitions.
Within training categories, object instances are separated into 80\%/10\%/10\% train/validation/test partitions for unseen-object evaluation.
The category and object assignments are disjoint at their stated level.
Unseen-instruction robustness is different by design: it evaluates the index-0 unseen wording on the same training object--part pairs used with index-1/2 instructions for optimization, isolating sensitivity to instruction wording rather than geometric transfer.

\begin{figure*}[t]
\centering
\begin{minipage}[t]{0.485\textwidth}
  \centering
  \includegraphics[width=\linewidth]{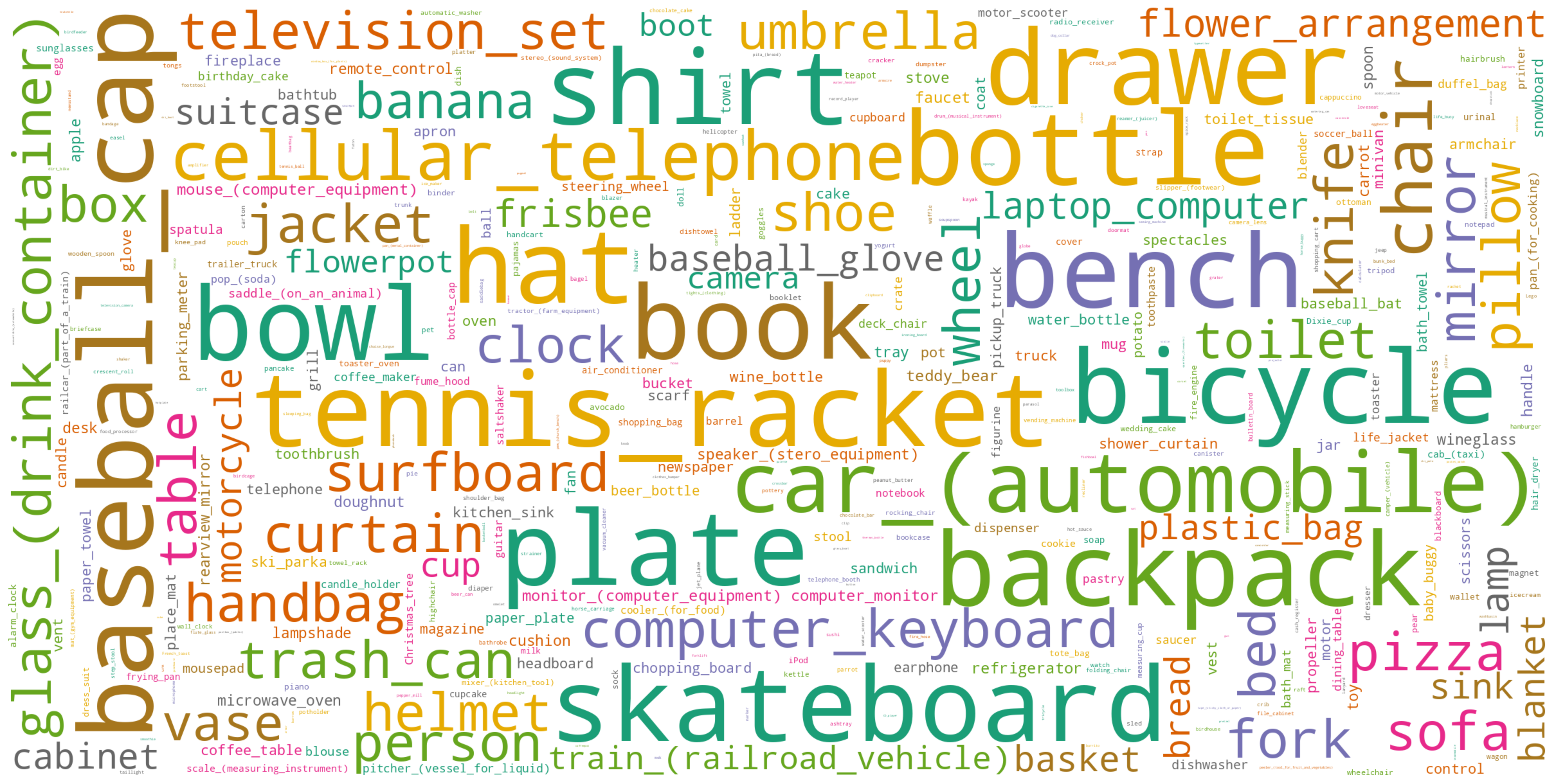}
  \par\smallskip
  \small (a) Object-category names.
\end{minipage}
\hfill
\begin{minipage}[t]{0.485\textwidth}
  \centering
  \includegraphics[width=\linewidth]{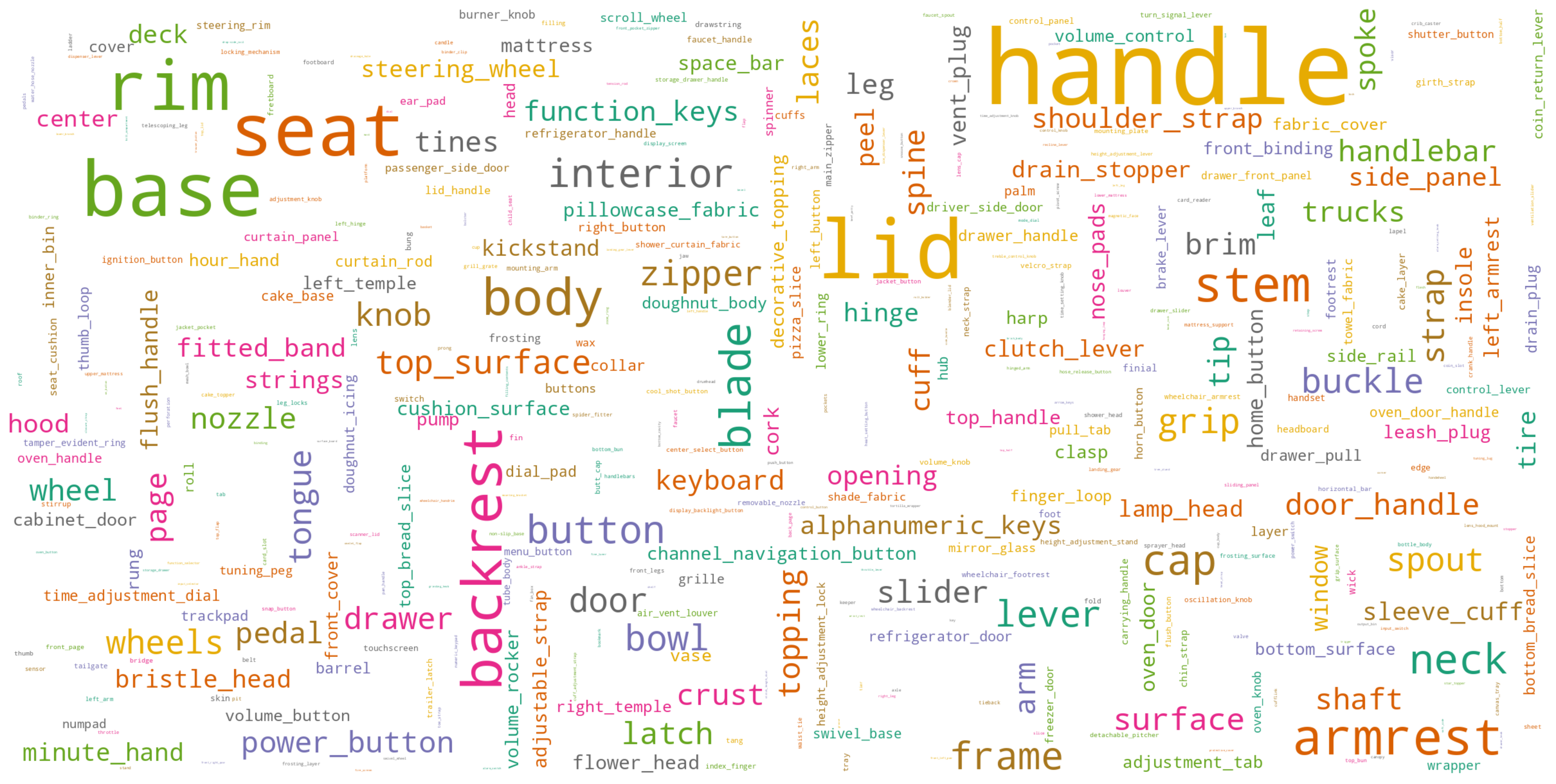}
  \par\smallskip
  \small (b) Normalized part names.
\end{minipage}
\caption{Dataset vocabulary word clouds. Label size is proportional to frequency in the word-cloud counts; colors separate neighboring labels and do not encode semantic groups. Underscores preserve multi-word canonical names. The visualization is descriptive: exact split sizes and unique-vocabulary counts are reported in Table~\ref{tab:dataset_splits_app}.}
\label{fig:dataset_wordclouds}
\end{figure*}

Figure~\ref{fig:dataset_wordclouds} complements the aggregate counts with a vocabulary-level view.
The dominant names reveal common object and interaction-part concepts, while the many small entries make the benchmark's long tail visible.

\section{Model and Optimization Details}
\label{app:impl_details}

\subsection{Frozen VLM Representation and Training Cache}

The frozen encoder is \texttt{nvidia/Cosmos-Reason2-2B} with hidden width $D=2{,}048$.
It receives one masked source image (white outside the object) and the instruction followed by the marker \texttt{<AFF>}.
The marker is added to the tokenizer once and initialized to the mean existing token embedding.
The VLM remains in evaluation mode: the marker embedding and VLM weights are not optimized during decoder training.
Instead, the contextual hidden state at the marker position is cached as $\mathbf{h}_{\mathrm{aff}}\in\mathbb{R}^{2048}$.
The source-image tokens are pooled to a $12\times12$ grid, giving $\mathbf{F}_v\in\mathbb{R}^{144\times2048}$.
Only the downstream projection, prototypes, fusion blocks, and prediction head are learned.

This separation resolves the cache semantics: ``instruction-conditioned'' refers to the cached contextual state produced by the frozen VLM, whereas ``learnable'' refers to decoder-side prototypes and projections.
Projection coordinates and visibility flags are cached with the features so every decoder run uses the same point-to-image correspondence.

\paragraph{Training and inference interface.}
The task input at inference is one RGB image, its known binary object mask, and a free-form instruction.
For training, SAM-3D-Objects geometry, camera projection metadata, and the frozen Cosmos states described above are computed offline and cached; only the decoder is optimized from these records.
At inference, the RGB--mask pair first passes through SAM-3D-Objects to obtain the object-centric Gaussian representation and source-camera correspondence.
The same source RGB, whitened outside the known mask, and the instruction are then encoded online by the frozen Cosmos model, after which the trained decoder predicts one affordance probability per reconstructed point.
Thus inference requires neither ground-truth 3D geometry nor a precomputed VLM cache.

\subsection{Decoder Tensor Flow}

\begin{table*}[t]
\centering
\small
\setlength{\tabcolsep}{5pt}
\begin{tabular}{lll}
\toprule
Tensor & Shape per sample & Role \\
\midrule
$\mathbf{X}$ & $4{,}096\times3$ & normalized Gaussian centers \\
$\mathbf{F}$ & $4{,}096\times13$ & cached point features \\
$\mathbf{h}_{\mathrm{aff}}$ & $2{,}048$ & instruction-conditioned global VLM state \\
$\mathbf{F}_v$ & $144\times2{,}048$ & $12\times12$ source-image token grid \\
$\mathbf{P}$ & $4{,}096\times256$ & encoded point tokens \\
$\mathbf{Z}$ & $16\times256$ & conditioned semantic prototypes \\
$\hat{\mathbf{s}}$ & $4{,}096$ & predicted point probabilities \\
\bottomrule
\end{tabular}
\caption{Principal tensors in the AffordAny decoder. Batch dimensions are omitted.}
\label{tab:tensor_shapes}
\end{table*}

The point encoder has three residual MLP blocks with width 256.
Local visual features are bilinearly sampled from the $12\times12$ token grid and injected into visible point tokens.
Semantic compression uses 16 prototypes, FiLM conditioning from $\mathbf{h}_{\mathrm{aff}}$, and a two-layer Transformer decoder.
One Group--Mix--Ungroup block exchanges information between prototypes and points; an eight-head, four-layer point-to-semantic Transformer cross-decoder then maps the fused representation to point features, matching the architecture described in the main paper.
Dropout is 0.1.
The prediction head is LayerNorm $\rightarrow$ Linear $\rightarrow$ GELU $\rightarrow$ Dropout $\rightarrow$ Linear$(1)$.
The decoder contains 10.54M trainable parameters, excluding the frozen VLM.

\subsection{Optimization}

The main AffordAny model and the full baseline comparison are trained for 300 epochs with seed 42.
We use AdamW, per-GPU batch size 16 on eight GPUs, automatic mixed precision, gradient clipping at 1.0, weight decay $5\times10^{-4}$, five warmup epochs, and cosine learning-rate decay from $2\times10^{-4}$ to $10^{-5}$.
The loss weights are $\lambda_{\mathrm{bce}}=0.2$, $\lambda_{\mathrm{focal}}=1.0$, and $\lambda_{\mathrm{dice}}=0.5$.
Weighted BCE uses positive weight 8.0; focal loss uses $\alpha=0.75$ and $\gamma=2.0$.
Instruction dropout is 0.2 for AffordAny and disabled at evaluation.
Checkpoints are evaluated every ten epochs and selected by unseen-object validation IoU; AffordAny's selected checkpoint is epoch 140.
The optimization settings are kept unchanged for the component ablation experiments.

\subsection{Pseudo-Label Self-Training}

The teacher processes 5{,}325 LVIS objects disjoint from the supervised benchmark, producing 32{,}718 object--part--instruction pseudo-label records (10{,}906 index-0 and 21{,}812 index-1/2 records).
With seed 42, a fixed 0.37 subsample retains 12{,}105 pseudo records, giving a near-1:1 mixture with the 12{,}134 real training records.
For each pseudo record, points with teacher probability in the closed interval $[0.15,0.65]$ are ignored, and the remaining predictions are binarized.
Every pseudo-label sample receives the fixed weight
\begin{equation}
\bar{w}=0.1.
\end{equation}
The self-training stage uses hard targets after the ignore mask, applying the supervised $q_{ij}>0$ rule.
It starts from the supervised checkpoint and runs on five GPUs for 100 epochs (batch size 16 per GPU, learning rate $5\times10^{-5}$, ten warmup epochs, and no mixed precision).
Epoch 1 and every tenth epoch are evaluated, and the checkpoint selected by unseen-object validation IoU is used for all reported self-training results.
Pseudo BCE and focal contributions receive an additional factor of 0.5, giving approximately 5\% of the corresponding real-sample weight, while the overall loss uses $(\lambda_{\mathrm{bce}},\lambda_{\mathrm{focal}},\lambda_{\mathrm{dice}})=(0.2,1.0,0.7)$.

For post-hoc calibration, validation logits fit the monotonic transform
\begin{equation}
\hat{s}^{\mathrm{cal}}_i=\sigma(1.10z_i-0.75),
\end{equation}
where $z_i$ is the raw self-trained logit.
This transform preserves ranking and therefore AUC, but it changes fixed-threshold IoU and MAE.
Raw and calibrated predictions are consequently reported separately.

\subsection{Computational Cost and Practical Feasibility}
\label{app:compute_cost}

We profile the exact epoch-140 decoder architecture with batch size 1, $N=4{,}096$ points, 144 visual tokens, 16 prototypes, and width 256.
Following PyTorch's operation-level profiler conventions, the reported 121.63 GFLOPs quantify matrix operations in the executed forward graph and comprise 69.14 GFLOPs of batched matrix multiplication and 52.46 GFLOPs of linear matrix multiplication.
Other operators, including softmax, normalization, GELU, indexing, and bilinear sampling, are reflected in the measured latency and memory results rather than the matrix-operation FLOP count.

\begin{table}[t]
\centering
\small
\setlength{\tabcolsep}{4pt}
\begin{tabular}{lr}
\toprule
Cached-decoder quantity & Value \\
\midrule
Trainable parameters & 10.54M \\
BF16 parameter footprint & 20.1 MiB \\
Profiled matrix operations & 121.63 GFLOPs \\
Attention batched matmul & 69.14 GFLOPs (56.8\%) \\
Linear matmul & 52.46 GFLOPs (43.1\%) \\
BF16 latency, median / P95 & 16.5 / 18.6 ms \\
Peak additional allocated memory & 45.3 MiB \\
\bottomrule
\end{tabular}
\caption{Batch-1 decoder cost for one 4{,}096-point object. Latency and allocated-memory measurements use BF16 autocast on one NVIDIA A100-SXM4-80GB after 20 warm-up iterations and over 100 timed forwards; inputs are already resident on the GPU.}
\label{tab:compute_cost}
\end{table}

The dominant term is the self-attention inside the four point-decoder layers.
Its attention-score and value products require approximately $4LN^2d=68.7$ GFLOPs for $L=4$, $N=4{,}096$, and $d=256$, accounting for nearly all attention batched-matmul cost.
In contrast, point--prototype interactions scale with $NKd$ for $K=16$.
The measured 16.5-ms median corresponds to about 60 cached-decoder evaluations per second on the stated GPU.
This measurement isolates the learned localization decoder under the cached-feature setting used for training and controlled comparison.

\section{Evaluation Protocol}
\label{app:protocol}

\subsection{Baseline Implementations}

All methods receive the same 4{,}096-point geometry, masks, labels, and data splits, and use the same losses, 300-epoch budget, seed, per-GPU batch size, and unseen-object validation criterion.
Their frozen language or vision--language features follow the corresponding method design rather than being forced to share one encoder.
OpenAD uses a pooled 512-dimensional CLIP text embedding and point--text cosine alignment.
LASO uses 77 CLIP text tokens, two affordance-fusion modules (AFMs, implemented as LASO GPBlocks), and a two-layer text-query RPD.
LMAffordance3D uses the same token count with two spatial-fusion layers and two cross-attention decoder layers.
AffordAny instead uses the cached Cosmos contextual state and 144 source-image tokens with 16 semantic prototypes, one GPBlock, and a four-layer decoder.

\begin{table*}[t]
\centering
\small
\setlength{\tabcolsep}{4pt}
\begin{tabular}{llllr}
\toprule
Method & Frozen feature input & Adapted decoder & Trainable params. & Selected epoch \\
\midrule
OpenAD & CLIP pooled text (512-D) & point--text cosine alignment & 6.31M & 130 \\
LASO & 77 CLIP text tokens & 2 AFM/GPBlocks + 2-layer RPD & 7.01M & 120 \\
LMAffordance3D & 77 CLIP text tokens & 2 fusion + 2 decoder layers & 6.38M & 140 \\
AffordAny & Cosmos state + 144 visual tokens & prototypes + GPBlock + 4-layer decoder & 10.54M & 140 \\
\bottomrule
\end{tabular}
\caption{Controlled decoder implementations used in the main comparison. Parameter counts exclude frozen feature encoders. The comparison standardizes data, supervision, optimization, and model selection while preserving each method's native feature representation and decoder design.}
\label{tab:baseline_implementations}
\end{table*}

\subsection{Metric Definitions}

Let $t_{ij}\in[0,1]$ be the packaged soft heatmap and let $p_{ij}\in[0,1]$ be the clamped predicted probability for point $j$ of evaluation row $i$.
The implementation forms the binary label $y_{ij}=\mathbb{1}[t_{ij}>0]$ (\texttt{positive\_threshold}$=0$) and prediction $\hat y_{ij}=\mathbb{1}[p_{ij}\ge0.5]$.
Unknown points have already been removed when the 4{,}096-point evaluation tensors are packaged, so all tensor entries are valid in these computations.
IoU pools intersections and unions over all rows:
\begin{equation}
\mathrm{IoU}=\frac{\sum_{ij}\hat y_{ij}y_{ij}}{\max\!\left(1,\sum_{ij}\mathbb{1}[\hat y_{ij}+y_{ij}>0]\right)}.
\end{equation}
For mIoU, intersections and unions are first pooled within each object category, each category denominator is likewise clamped to at least one, and the resulting category IoUs are averaged, giving equal weight to head and tail categories.
AUC is computed globally over all points with the rank-sum form of ROC AUC (and is undefined if either class is absent), while
\begin{equation}
\mathrm{MAE}=\frac{1}{\sum_i N_i}\sum_{ij}\lvert p_{ij}-y_{ij}\rvert
\end{equation}
uses the binary labels rather than the soft heatmap.
SIM instead retains the soft target: for each row, $p_i$ and $t_i$ are independently normalized to unit mass and their histogram intersection $\sum_j\min(\bar p_{ij},\bar t_{ij})$ is computed, then averaged over rows.
The implementation assigns SIM 1 if both row sums are zero and 0 if exactly one is zero.
IoU and mIoU are threshold dependent; AUC and SIM provide complementary ranking and distributional views.

\subsection{Human-Verified Evaluation Subsets}
\label{app:human_review}

We additionally constructed human-verified evaluation subsets from the test samples.
Each row was manually verified using the masked source image and six canonical renders.
A row was retained only when the reconstruction was usable, the target part was present and unambiguous, the part localization was accurate, and the instruction was natural and visually groundable.
The final human-verified subsets contain 41 unseen-category rows and 54 unseen-object rows, for 95 rows in total, and provide human-validated ground truth for independently assessing model performance.

For consistent dense evaluation, labels marked \emph{ignore} are conservatively mapped to background.
All four frozen checkpoints are evaluated at threshold 0.5 with the same pooled IoU, category-macro mIoU, global ROC AUC, and pointwise MAE definitions.
Table~\ref{tab:human_reviewed_diagnostic} reports performance on the two human-verified subsets.

\begin{table*}[t]
\centering
\small
\setlength{\tabcolsep}{4pt}
\begin{tabular}{lllccccc}
\toprule
Analysis & Split & Model & $N$ & IoU $\uparrow$ & mIoU $\uparrow$ & AUC $\uparrow$ & MAE $\downarrow$ \\
\midrule
\multirow{4}{*}{Human-verified}
 & \multirow{4}{*}{Unseen category}
 & \textbf{AffordAny} & 41 & \textbf{.3387} & \textbf{.2537} & \textbf{.7095} & \textbf{.3030} \\
 & & LASO & 41 & .2628 & .1953 & .6879 & .3193 \\
 & & LMAffordance3D & 41 & .2596 & .1988 & .6452 & .3151 \\
 & & OpenAD & 41 & .2897 & .2437 & .6747 & .3715 \\
\midrule
\multirow{4}{*}{Human-verified}
 & \multirow{4}{*}{Unseen object}
 & \textbf{AffordAny} & 54 & \textbf{.4371} & \textbf{.3570} & \textbf{.8556} & .1901 \\
 & & LASO & 54 & .3656 & .2766 & .8299 & .2165 \\
 & & LMAffordance3D & 54 & .4133 & .3299 & .8409 & \textbf{.1787} \\
 & & OpenAD & 54 & .3360 & .2653 & .7902 & .2846 \\
\bottomrule
\end{tabular}
\caption{Evaluation on the human-verified subsets, containing 41 unseen-category rows and 54 unseen-object rows. Bold denotes the best value within each split.}
\label{tab:human_reviewed_diagnostic}
\end{table*}

\subsection{Paired Statistical Tests}

Self-training comparisons reuse matched evaluation rows and run a paired bootstrap with 10{,}000 resamples and seed 42.
Each resample draws evaluation rows with replacement and recomputes pooled IoU, category-macro mIoU, global ROC AUC, and pointwise MAE.
Percentile intervals give the reported 95\% confidence bounds, and the corresponding paired bootstrap distributions provide the significance markers reported in the main paper.
To preserve matched comparisons, resampling is performed at the evaluation-row level under the fixed evaluation manifest and checkpoints.

\begin{table}[t]
\centering
\small
\setlength{\tabcolsep}{3pt}
\begin{tabular}{llrr}
\toprule
Split & Metric & Raw $\Delta$ & 95\% CI \\
\midrule
Unseen obj. & IoU  & $+0.0002$ & $[-0.0039,0.0042]$ \\
            & mIoU & $+0.0089$ & $[0.0050,0.0118]$ \\
Unseen cat. & IoU  & $+0.0098$ & $[0.0068,0.0131]$ \\
            & mIoU & $+0.0152$ & $[0.0104,0.0183]$ \\
\bottomrule
\end{tabular}
\caption{Paired-bootstrap differences for raw self-training minus the supervised baseline. For compactness, the table lists the overlap metrics; AUC and MAE are computed from the same paired resamples and their significance markers are reported in the main paper. Self-training significantly improves category-macro mIoU on both splits and IoU on unseen categories ($p<0.001$), while maintaining unseen-object IoU ($p=0.461$).}
\label{tab:bootstrap_ci}
\end{table}

\section{Extended Quantitative Analysis}
\label{app:quantitative}

\subsection{Per-Category Distribution}
\label{app:per_category}

\begin{figure*}[t]
\centering
\includegraphics[width=0.92\textwidth]{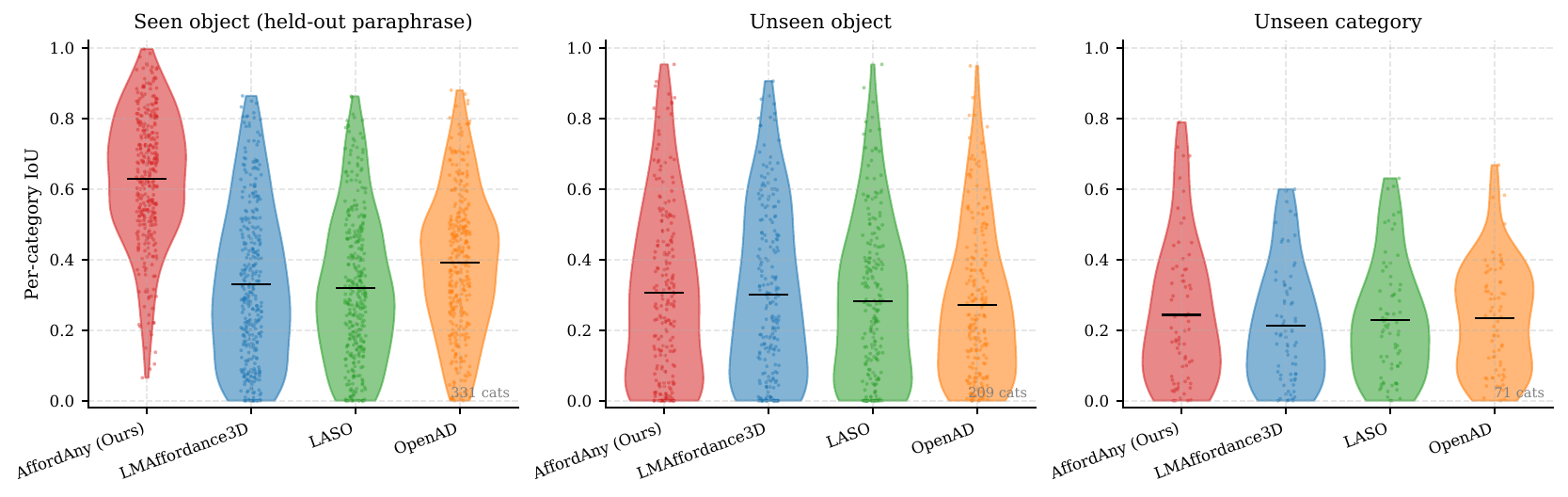}
\caption{Per-category IoU distributions. Violins summarize categories; dots are individual category values and black bars denote means. ``Unseen instruction'' evaluates unseen wording on the training objects.}
\label{fig:supp_a1}
\end{figure*}

Figure~\ref{fig:supp_a1} plots category-level IoU, where intersection and union are pooled within each category before plotting.
This avoids conflating categories with different numbers of points or rows.
AffordAny achieves the highest central tendency on unseen instructions and unseen categories across heterogeneous category distributions.

\section{Complementary Metric Analysis}
\label{app:metric_diagnostics}

IoU, AUC, MAE, and SIM provide complementary views of probabilistic point predictions.
At the fixed operating threshold of 0.5, IoU measures overlap between the selected 3D region and the target.
AUC evaluates threshold-free point-wise ranking, MAE summarizes probability error over positive and negative points, and SIM measures normalized distributional agreement.
This complementarity is particularly informative for sparse affordance regions, where calibration can move scores across the operating threshold without changing their ranking.
We therefore use IoU as the primary localization metric and jointly report AUC, MAE, and SIM for a broader characterization of prediction quality.

\subsection{Ranking--Overlap Discrepancy}

We inspect unseen-category test rows satisfying the direct constraints
\begin{equation}
\mathrm{AUC}>0.93,\qquad \mathrm{MAE}<0.015,\qquad \mathrm{IoU}\leq0.02.
\end{equation}
These constraints identify cases in which strong point ordering and low average probability error coexist with limited fixed-threshold overlap.
Unlike a percentile-based composite score, this transparent rule directly isolates the metric relationship of interest.

\begin{figure}[H]
  \centering
  \includegraphics[width=\columnwidth]{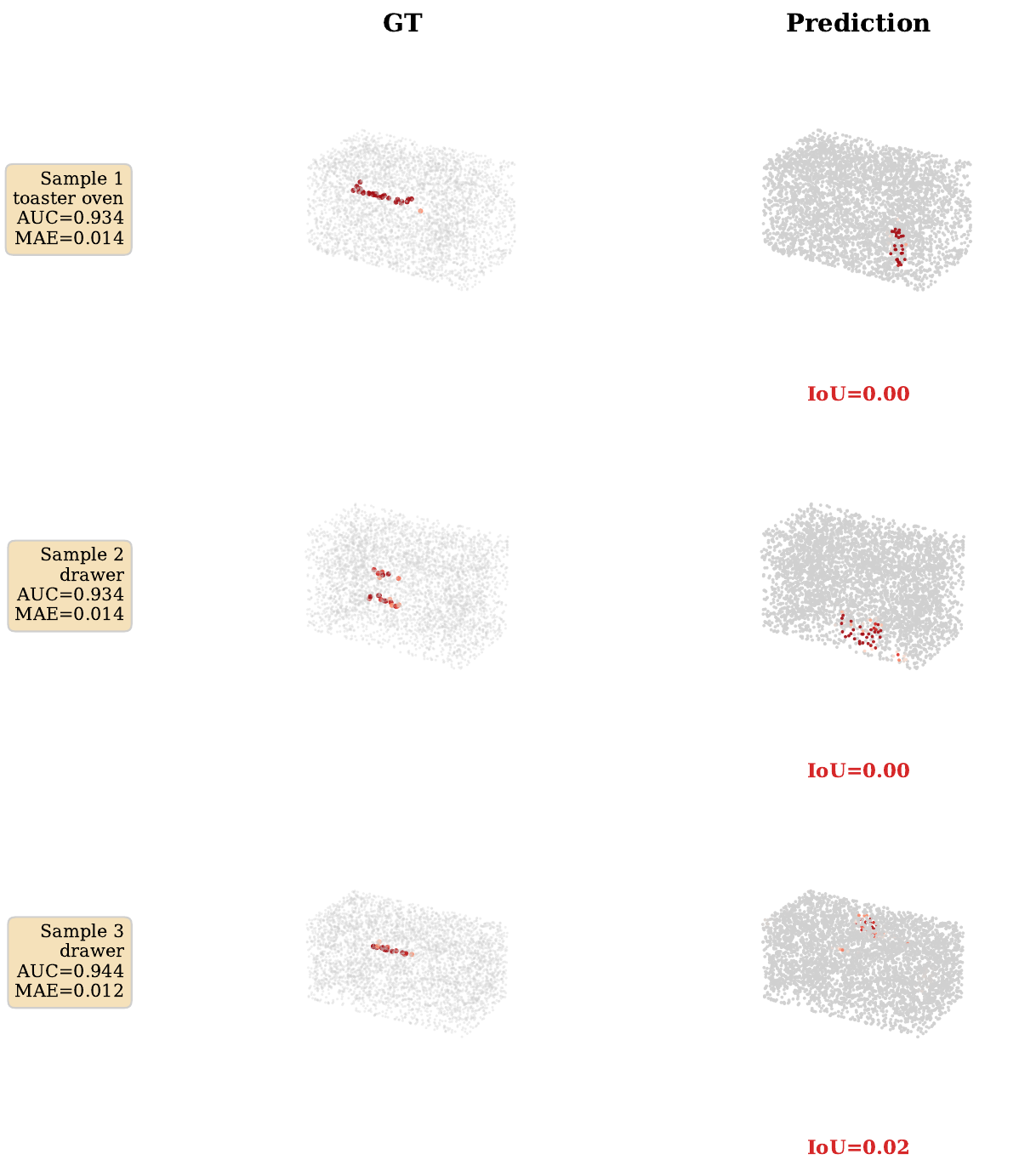}
  \caption{Metric-complementarity examples on unseen categories. Each block shows ground truth and the prediction at threshold 0.5. The values motivate joint reporting of overlap, ranking, and probability-error metrics.}
  \label{fig:supp_metric_discrepancy}
\end{figure}

Figure~\ref{fig:supp_metric_discrepancy} shows three such rows.
The toaster-oven example has IoU 0.000, AUC 0.934, and MAE 0.014; the first drawer has the same rounded metrics; and the second drawer has IoU 0.020, AUC 0.944, and MAE 0.012.
In each case, target points can rank above background points, yielding high AUC, while relatively few target scores exceed the 0.5 operating threshold.
MAE remains low because the majority of points are background.
Together, these examples show why ranking-based and average-error metrics alone do not fully characterize actionable localization.

\subsection{Operational Interpretation}

High AUC with low fixed-threshold IoU can arise when correctly ranked target points remain below 0.5 or when only a subset of the target is ranked above background.
The former primarily reflects calibration at the operating threshold, whereas the latter reflects localization coverage.
Reporting both thresholded and threshold-free metrics makes these effects explicit.
For deployment, the threshold should be selected on validation data and held fixed on the test set.
Within that protocol, IoU directly measures overlap of the actionable region, AUC measures ranking robustness, MAE measures probabilistic error, and SIM measures normalized distributional agreement.

\section{Qualitative Results and Challenging Cases}
\label{app:qualitative}

\begin{figure*}[t]
\centering
\includegraphics[width=0.86\textwidth]{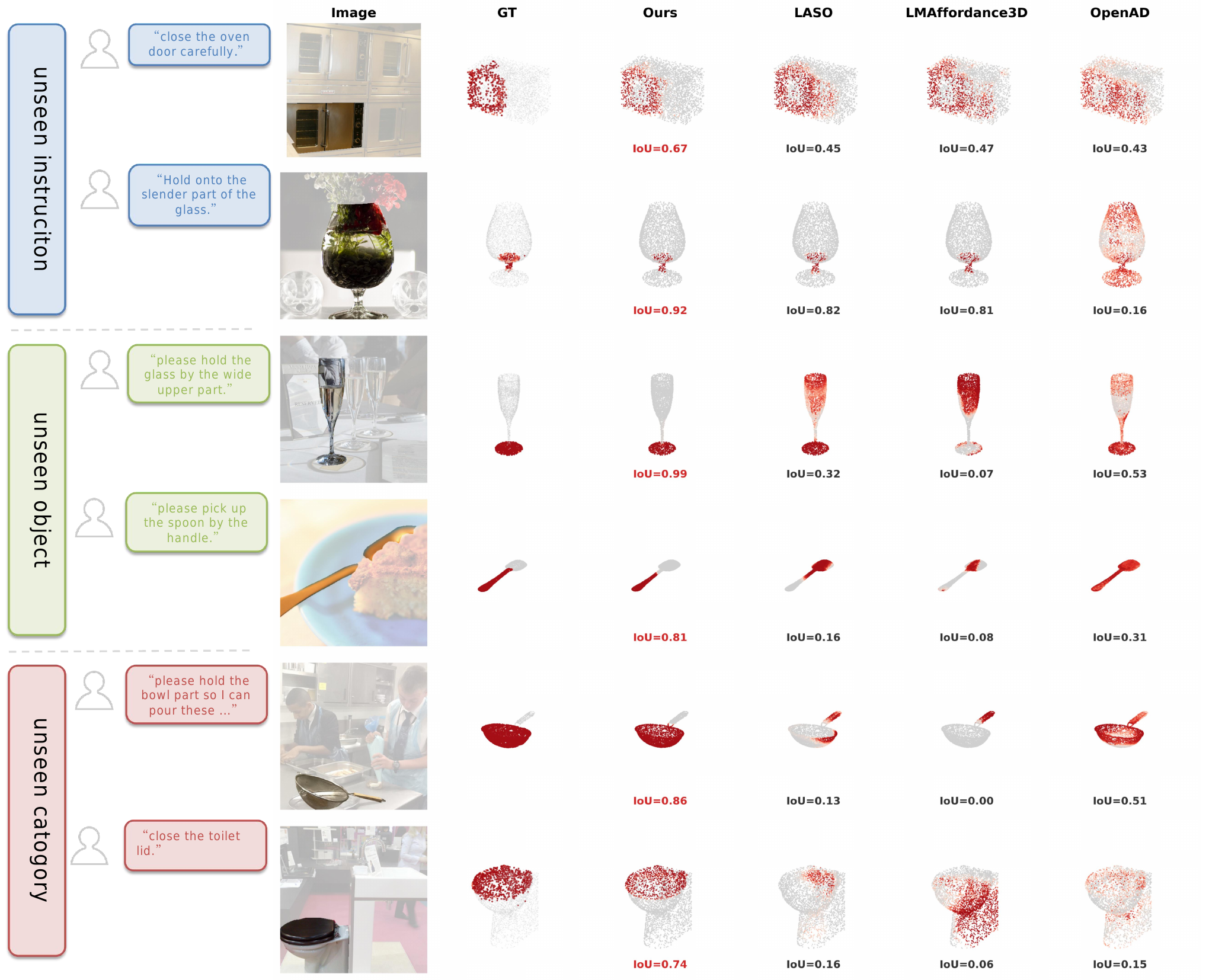}
\caption{Qualitative comparison across unseen instructions, unseen objects, and unseen categories. Each group shows the source image/instruction, ground truth, and predictions. Red points denote the thresholded affordance region and the numbers are per-row IoU. These examples are selected for clear visualization and complement the aggregate results reported in the main tables.}
\label{fig:qualitative_app}
\end{figure*}

Figure~\ref{fig:qualitative_app} compares matched rows from the three generalization conditions.
AffordAny generally produces a more concentrated region on the intended functional part, including thin stems and handles.
The examples highlight diverse object geometries and complement the full-split quantitative tables and category distributions.

\subsection{Robustness Considerations}

Performance is most challenging when uncertainty arises before or during decoding.
\textbf{Reconstruction quality} can be affected for thin, reflective, transparent, or heavily occluded parts.
\textbf{Canonical-view coverage} can be limited when a valid part is absent from all rendered views.
\textbf{Semantic proposals} can be less applicable when a category-level part is not present in a particular image.
\textbf{2D-to-3D lifting} is most sensitive to depth ownership and closely adjacent boundaries.
\textbf{Instruction grounding} is most challenging for related functional parts or spatially broad target regions.
Ambiguous commands may admit multiple valid targets.

Unknown masks reduce the effect of missing evidence by excluding uncertain regions from supervision.
\FloatBarrier

\section{Limitations and Release Constraints}
\label{app:limitations}

The benchmark is intended for research on language-conditioned object-part localization and open-world 3D affordance perception.
The current evaluation focuses on perception and grounding; integration with grasp planning, collision-aware motion, and physical execution is left for future work.
Single-view reconstruction may be less reliable on unobserved surfaces, and label quality remains dependent on upstream reconstruction and segmentation.
The benchmark reflects the naturally long-tailed distribution of object interactions, and its LVIS provenance motivates future expansion toward broader geographic and cultural coverage.

Release artifacts should preserve LVIS identifiers and provenance and must follow the applicable LVIS and source-image terms.
Generated annotations remain subject to upstream image redistribution rights.
The construction pipeline uses public benchmark imagery and introduces no private user interaction data.
The separate code/data package supports result reproduction through preprocessing scripts, frozen split manifests, configuration snapshots, checkpoint hashes, and commands for the reported evaluations.

\end{document}